\pdfoutput=1

\documentclass[11pt]{article}

\usepackage[]{acl} %
\usepackage{float}
\usepackage{times}
\usepackage{latexsym}

\usepackage[T1]{fontenc}

\usepackage[utf8]{inputenc}

\usepackage{microtype}

\usepackage{inconsolata}

\usepackage{graphicx}

\usepackage{multirow}
\usepackage{multicol}

\title{Convolutional Neural Networks can achieve binary bail judgement classification}

\author{Amit Barman$^{1\dagger}$, Devangan Roy $^{2\dagger}$, Debapriya Paul$^{3\ddag}\thanks{\hspace*{0.5em} The work was carried out when the author was in Jadavpur University.}$, \\ {\bf Indranil Dutta$^{4\dagger}$, Shouvik Kumar Guha$^{5\diamond}$, Samir Karmakar$^{6\dagger}$, Sudip Kumar Naskar$^{7\dagger}$} \\
        $^{\dagger}$Jadavpur University, Kolkata, India \\ 
        $^{\ddag}$Indian Institute of Engineering Science and Technology, Shibpur, India\\ 
        $^{\diamond}$ West Bengal National University of Juridical Sciences, Kolkata, India\\
        \{$^{1}$amitbarman811, $^{7}$sudip.naskar\}@gmail.com,\\ 
        \{$^{2}$devanganr.sll.rs, $^{4}$indranildutta.lnl , $^{6}$samir.karmakar \}@jadavpuruniversity.in, \\
        $^{3}$debapriya.rs2023@cs.iiests.ac.in, $^{5}$shouvikkumarguha@nujs.edu } 
\begin{document}

 \maketitle

\begin{abstract}


There is an evident lack of implementation of Machine Learning (ML) in the legal domain in India, and any research that does take place in this domain is usually based on data from the higher courts of law and works with English data. The lower courts and data from the different regional languages of India are often overlooked. In this paper, we deploy a Convolutional Neural Network (CNN) architecture on a corpus of Hindi legal documents. We perform a bail Prediction task with the help of a CNN model and achieve an overall accuracy of 93\% which is an improvement on the benchmark accuracy, set by \citet{kapoor-etal-2022-hldc}, albeit in data from 20 districts of the Indian state of Uttar Pradesh.

\end{abstract}

\section{Natural Language processing in the legal domain}
District and subordinate courts in India have a preponderance of denied bail pleas. In addition, Indian courts persistently have close to 40 million cases in pendency \cite{national-judicial-data-grid2021}.  In this paper, we report on the use of a convolutional neural network (CNN) architecture to train on judgments from the Hindi Legal Documents Corpus (HLDC) and predict the outcome of bail pleas into two categories - bail granted or bail denied.

In a seminal article, pretty much at the early onset of machine intelligence, \cite{berman1989} highlights the importance of the need to employ artificial intelligence, albeit in the context of the American legal system, to address crises that are both processual and financial that they observe are responsible for the lack of confidence that citizens have in the legal system. They propose a diagnostic model for sentencing that is true to the technology of the time decision tree. In recent times, \cite{medvedeva2018,Medvedeva2023} have employed, with a great deal of caution, use cases where they outline the need to look into three key areas of the use of artificial intelligence in law, namely, outcome identification, outcome-based
judgement categorization and outcome forecasting. What emerges in the literature are two crucial threads; caution in the use of artificial intelligence in Law \cite{Cofone2022} and the qualitative importance of training data in building reasonably well-argued outcomes \cite{medvedeva2018}.

In the specific context of the use of NLP in the legal domain, there are arguably two issues: language resources consist of data from higher courts and they tend almost exclusively to be in English (but see \citet{chalkidis2021}). This is as true globally, as it is in India, where the challenges are compounded due to the use of regional languages and the lack of development of NLP tools specific to the legal domain. In this paper, we report on successfully improving upon the benchmark prediction of Hindi bail plea outcomes as established in \citet{kapoor-etal-2022-hldc}(Hindi Legal Documents Corpus (HLDC)). Given the high number of cases in certain districts of Uttar Pradesh (UP), India compared to others, we deploy our CNN models on a set of 10 districts with higher numbers of cases and another set of 10 districts with lower number of cases. In the following section, we provide a brief description of the HLDC and the pre-processing routines followed by us. Section 3 details the experiments and the results. In section 4, we briefly summarize our results.

\section{Data}

\begin{figure*}[h!]
    \centering
    \includegraphics
    [scale=0.38]
    {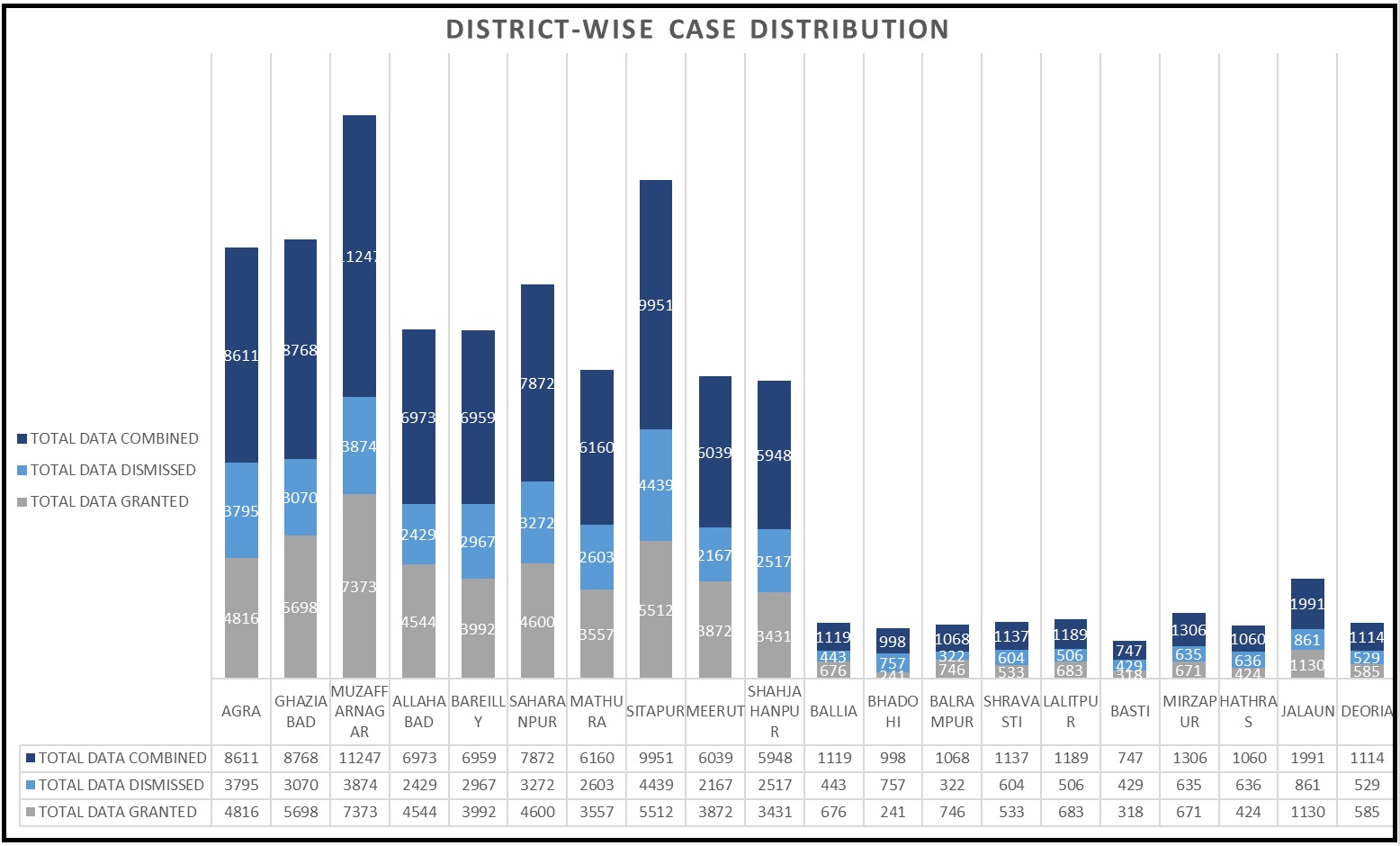}
    \caption{Bar plot showing the distribution of "granted" and "dismissed" data for the 20 selected districts}
    \label{fig-2}
\end{figure*}

The HLDC corpus was created by collecting data from the e-Courts website \citet{kapoor-etal-2022-hldc}. This corpus contains judgments from district courts of 71 districts in the state of Uttar Pradesh for a duration of 2 years from May 01, 2019, to May 01, 2021. We have used this corpus for training and testing our CNN. An obvious issue that arose post-pre-processing was the imbalance in several ``granted'' and ``dismissed'' cases for each district. Figure \ref{fig-2} shows the district-wise case distribution for the 20 selected districts. We can see that in nearly all the districts (excepting \emph{Bhadohi}, \emph{Shravasti}, \emph{Basti}, and \emph{Hathras}) the number of ``granted'' case data exceeds the number of ``dismissed'' case data and often by quite a lot. It is only for \emph{Deoria} and \emph{Mirzapur} that we see a nearly equal distribution of case data in terms of decision.

\subsection{Data pre-processing}


We begin our objective of pre-processing the data by first extracting the data from the HLDC dataset which was kindly provided to us upon request by \citet{kapoor-etal-2022-hldc}. We selected the ``processed'' data from the district court with the highest number of ``processed'' case data from each district file. We selected 20 districts based on the file size from all the districts available (10 districts with the highest file size and 10 districts with the lowest file size, we assumed file size as a parameter to define data load, as in, higher the file size higher the number of case data). We noticed that the data contained 3 types of values in the ``decision'' column - ``granted'', ``dismissed'', and ``don't know''. Upon further examination of the data, we discovered that the case data classified as ``don't know'' are actually ``granted'' and ``dismissed'' case data which has been misclassified. We also found that in quite a few instances, for all the districts, some of the case data classified as ``granted'' under ``decision'' has a ``bail\textunderscore amount'' of ``-1'' or ``0'' assigned to it, while some which are classified as ``dismissed'' has a ``bail\textunderscore amount'' other than ``-1'' or ``0'', like ``20000'' or ``25000''. A ``granted'' bail decision should be accompanied by a specific bail amount like ``15000'' or ``50000'' and not ``-1'' or ``0''. Similarly, a ``dismissed'' bail decision should not be accompanied by a bail amount, and thus should have values like ``-1'' or ``0'' in ``bail\textunderscore amount''. With this knowledge we sieved out (1) all the case data which had ``don't know'' as the ``decision'', (2) all the case data with ``dismissed'' in ``decision'' but any other ``bail\textunderscore amount'' than ``-1'' or ``0'', (3) all the case data with ``granted'' in ``decision'' but ``-1'' or ``0'' in ``bail\textunderscore amount'' from the files for each of the selected districts. With this our pre-processing of the data was complete. 

We split our data after pre-processing into an 80:20 where the bulk part (i.e. 80\% of the data) is used for training and validation purposes (we use the same data for training and validation) and the remaining 20\% of the data forms our test set. Table \ref{tab-1} shows the training and test set data for each of the 20 selected districts and for the total dataset as well as the final data distribution of our dataset after pre-processing. \footnote{Anonymized pre-processed data and the code will be made available on https://github.com/indranildutta/JU-NUJS.}

\begin{table}[h!]\small
\centering
\begin{tabular}{rccc}
\hline

\textbf{District} & \textbf{Total} & \textbf{Train} & \textbf{Test}\\
\hline
Agra & 8611& 6893 & 1718 \\
Allahabad & 6973 & 5578 & 1395 \\
Ballia & 1119 & 895 & 224 \\
Balrampur & 1068 & 854 & 214 \\
Bareilly & 6959 & 5567 & 1392 \\
Basti & 747 & 597 & 150 \\
Bhadohi & 998 & 798 & 200 \\
Deoria & 1114 & 891 & 223 \\
Ghaziabad & 8768 & 7014 & 1754 \\
Hathras & 1060 & 848 & 212 \\
Jalaun & 1991 & 1592 & 399 \\
Lalitpur & 1189 & 951 & 238 \\
Mathura & 6160 & 4928 & 1232 \\
Meerut & 6039 & 4831 & 1208 \\
Mirzapur & 1306 & 1044 & 262 \\
Muzaffarnagar & 11247 & 8997 & 2250 \\
Saharanpur & 7872 & 6297 & 1575 \\
Shahjahanpur & 5948 & 4758 & 1190 \\
Shravasti & 1137 & 909 & 228 \\
Sitapur & 9951 & 7960 & 1991 \\
\hline
\textbf{Total} & \textbf{90257} & \textbf{72202} & \textbf{18055} \\
\hline
\end{tabular}
\caption{District-wise data distribution for the 20 selected districts} 
\label{tab-1}
\end{table}

\section{Experiments and Results}
\begin{figure*}[h!]\small
    \centering
    \includegraphics
    [scale=0.75]
    {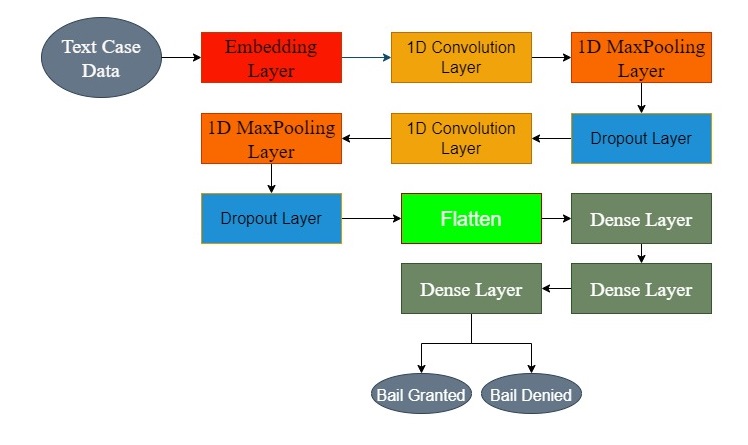}
    \caption{The CNN-text classifier for Bail Prediction}
    \label{fig-1}
\end{figure*}

It has been demonstrated that Convolution Neural Networks (CNNs) are effective tools for image analytics, particularly when used in conjunction with transfer learning to extract features \cite{Lu2020,he2016,Shawn2016,Bhangale2023}. In the past few years, CNN has been modified for text classification and has shown effectiveness in tasks like classification \cite{Shah2023,Minaee2021} where we anticipate finding strong local cues about class membership like a few salient lines or phrases \cite{lecun1998}. Convolution layers in a CNN for text classification use max pooling to condense or summarize the features that are extracted from the convolution and one-dimensional convolution with a small size kernel to extract features. Ultimately, the fully connected layer creates predictions by fitting the characteristics obtained from activations to the training set. We have also used this architecture in our paper for the purpose of our mission. Figure \ref{fig-1} displays the comprehensive architecture.
Word representations are the neural network's inputs, regardless of the neural network architecture that is selected. Word embedding is frequently used in neural networks for text classification to carry out the task. The process of word embedding involves treating every word as a vector in a low-dimensional space. 


Any text classification task involving deep learning must include tokenization; for our research, we have utilized hugging-face's ``bert-base-multilingual-uncased'' as our tokenizer \citep{devlin_2018}. which contains 110,000 words in its vocabulary. The embedding layer receives these tokens that were produced during the tokenization process.


\subsection{Model Architecture}
We employ a common architecture in this research, which comprises a fully connected model, a convolutional model, and a word embedding. Our model is implemented using Keras, a Tensorflow backend \cite{tensorflow2015-whitepaper}. It consists of two convolutional layers, the first of which has 128 filters and the second of which has 256 filters. The layer with an increasing number of filters is what makes up our Convolutional Neural Network classifier. Following each convolutional layer with a kernel size of five and an activation function that uses the Rectified Linear unit, there is one 1D Maxpooling layer, a Dropout layer with a dropoutout rate of 0.3, and each layer. The resultant arrays from the pooled feature maps are then flattened to produce a single continuous linear vector.
The completely linked layer, which is made up of three dense layers, then receives the flattened matrix as input. The first dense layer consists of 128 units, the second layer of 64 units, and the last layer of one unit of dense layer.  Except for the last layer, which has a sigmoid function as an activation mechanism, each of them has a rectified linear function as an activation function. There is a dropout layer with a dropout rate of 0.3 after the first dense layer. The purpose of adding this dropout layer is to stop the model from overfitting.

\subsection{Experiment modelling}
We combined data from 10 districts with relatively high case numbers and 10 districts with relatively low case numbers to test our model before testing all the data to ensure the robustness and sparsity of our model architecture in our research. and also independently ran our model in each of the 20 districts. Without altering the fundamental architecture, the padding in each case has been changed based on the longest scenario that can exist. You can view all of the results for the districts with comparatively high case numbers in the figure below.

\begin{table}[h!]\small
\centering
\begin{tabular}{rcc}
\hline
\textbf{District} & \textbf{Accuracy} & \textbf{Macro F1}\\
\hline
\multicolumn{3}{l}{Highest number of case documents}\\
\hline
Agra & 0.96 & 0.96 \\
Allahabad & 0.93 & 0.92 \\
Bareilly & 0.94 & 0.94 \\
Ghaziabad & 0.96 & 0.95 \\
Mathura & 0.89 & 0.89 \\
Meerut & 0.93 & 0.92 \\
Muzaffarnagar & 0.95 & 0.95 \\
Saharanpur & 0.95 & 0.95 \\
Shahjahanpur & 0.95 & 0.94 \\
Sitapur & 0.95 & 0.95 \\
\hline
\multicolumn{3}{l}{Lowest number of case documents}\\
\hline
Ballia & 0.94 & 0.93 \\
Balrampur & 0.88 & 0.86 \\
Basti & 0.88 & 0.88 \\
Bhadohi & 0.88 & 0.82 \\
Deoria & 0.91 & 0.91 \\
Hathras & 0.88 & 0.88 \\
Jalaun & 0.92 & 0.92 \\
Lalitpur & 0.90 & 0.90 \\
Mirzapur & 0.82 & 0.82 \\
Shravasti & 0.84 & 0.84 \\
\hline
\end{tabular}
\caption{District-wise Accuracy for the 20 districts}
\label{tab-5}
\end{table}

\subsubsection{High case numbers}
Our model has allowed us to reach the highest accuracy of 0.96 on test data where the number of cases is relatively higher. Table \ref{tab-2} shows the confusion matrix of the test.

\begin{table}[h!]\small
\centering
\begin{tabular}{rcc}
\hline
\textbf{} & \textbf{Granted} & \textbf{Dismissed} \\
\hline
\textbf{Granted} & 5887 & 340 \\
\textbf{Dismissed} & 293 & 9186 \\
\hline
\textbf{Precision} & 0.95 & 0.96 \\
\textbf{Recall} & 0.95 & 0.97 \\
\textbf{F1 Score} & 0.95 & 0.97 \\
\hline
\textbf{Accuracy} &  \multicolumn{2}{c}{\textbf{0.96}}\\ 
\hline
\end{tabular}
\caption{\label{tab-2} Confusion matrix for the 10 districts (collected) with the highest number of cases}
\end{table}


\subsubsection{Low case numbers}
Similarly, we tested our model in districts with comparatively lower case counts than in the higher number of case districts. Using test data, we are able to achieve an accuracy of 0.93. Table \ref{tab-3} displays the confusion matrix below.

\begin{table}[h!]\small
\centering
\begin{tabular}{rcc}
\hline
\textbf{} & \textbf{Granted} & \textbf{Dismissed} \\
\hline
\textbf{Granted} & 1060 & 84 \\
\textbf{Dismissed} & 83 & 1119 \\
\hline
\textbf{Precision} & 0.93 & 0.93 \\
\textbf{Recall} & 0.93 & 0.93 \\
\textbf{F1 Score} & 0.93 & 0.93 \\
\hline
\textbf{Accuracy} &  \multicolumn{2}{c}{\textbf{0.93}}\\ 
\hline
\end{tabular}
\caption{\label{tab-3} Confusion matrix for the 10 districts (collected) with lowest data}
\end{table}






\subsubsection{Results}
Our model outperformed the benchmark accuracy of 0.82 and produced better results in both low and high-case numbers. We conducted our experiment using the same model architecture across all 20 districts to obtain our final result. And employing the CNN-text classifier, we were able to obtain an accuracy of 0.93 on the test data. Table \ref{tab-4} contains the confusion matrix of the outcome below and Table \ref{tab-5} shows the comparison of our result with that of \citet{kapoor-etal-2022-hldc}.

\begin{table}[h!]\small
\centering
\begin{tabular}{rcc}
\hline
\textbf{} & \textbf{Granted} & \textbf{Dismissed} \\
\hline
\textbf{Granted} & 7285 & 424 \\
\textbf{Dismissed} & 927 & 9416 \\
\hline
\textbf{Precision} & 0.89 & 0.96 \\
\textbf{Recall} & 0.94 & 0.91 \\
\textbf{F1 Score} & 0.92 & 0.93 \\
\hline
\textbf{Accuracy} &  \multicolumn{2}{c}{\textbf{0.93}}\\ 
\hline
\end{tabular}
\caption{\label{tab-4} Confusion matrix for all 20 districts' collected data}
\end{table}


\begin{table}[h!]\small
\centering
\begin{tabular}{rrc}
\hline
\textbf{Paper} & \textbf{Model} & \textbf{Accuracy}\\
\hline
\textcolor{black}{Kapoor et al. (2022)} & TF-IDF+IndicBert & 0.82 \\
This paper & CNN Model & 0.93 \\ 
\hline
\end{tabular}
\caption{\label{tab-5} Result comparison}
\end{table}


\section{Conclusion}

In India, most of the research in the legal domain involving Machine Learning methods is performed on English language data from the Higher courts of law \cite{malik2021-ildc,strickson2020,zhong2019}. We have shown that ML can also be applied to data from the lower courts, and the district courts, in the regional languages to yield successful results. In this paper, we report on the use of convolutional layers to achieve a binary classification of legal data in Hindi. We apply the CNN architecture to the dataset that we extract from the Hindi Legal Documents Corpus (HLDC) via pre-processing. We then use this model to perform the binary classification task of Bail Prediction. Using the CNN model, we improve on the prediction task's benchmark accuracy set by \citet{kapoor-etal-2022-hldc}.

\section*{Acknowledgements}
We would like to acknowledge our gratitude towards the West Bengal National University of Juridical Sciences for extending financial support for conducting this research.

\bibliography{anthology,custom,reference}

\end{document}